\documentclass[runningheads]{llncs}
\usepackage{amsmath}
\usepackage{graphicx}
\usepackage{float}
\usepackage{caption}
\usepackage{subcaption}
\usepackage{xcolor}
\usepackage[export]{adjustbox}
\usepackage{overpic}
\usepackage{hyperref}
\hypersetup{
  colorlinks   = true, 
  urlcolor     = magenta, 
  linkcolor    = black, 
  citecolor    = black 
}
\begin{document}
\title{Virtual Home Staging:\\ Inverse Rendering and Editing an Indoor Panorama under Natural Illumination}

\titlerunning{Virtual Home Staging}
%
\author{Guanzhou Ji\inst{1}\orcidID{0000-0002-0673-3286} 
\and Azadeh O. Sawyer\inst{1}\orcidID{0000-0003-1750-5892} 
\and Srinivasa G. Narasimhan\inst{1}\orcidID{0000-0003-0389-1921}}
\authorrunning{G. Ji et al.}
%
\institute{Carnegie Mellon University, Pittsburgh PA 15213, USA\\
\email{\{gji,asawyer,srinivas\}@andrew.cmu.edu}}
\maketitle 
\begin{abstract} We propose a novel inverse rendering method that enables the transformation of existing indoor panoramas with new indoor furniture layouts under natural illumination. To achieve this, we captured indoor HDR panoramas along with real-time outdoor hemispherical HDR photographs. Indoor and outdoor HDR images were linearly calibrated with measured absolute luminance values for accurate scene relighting. Our method consists of three key components: (1) panoramic furniture detection and removal, (2) automatic floor layout design, and (3) global rendering with scene geometry, new furniture objects, and a real-time outdoor photograph. We demonstrate the effectiveness of our workflow in rendering indoor scenes under different outdoor illumination conditions. Additionally, we contribute a new calibrated HDR (Cali-HDR) dataset that consists of 137 calibrated indoor panoramas and their associated outdoor photographs.

\end{abstract}
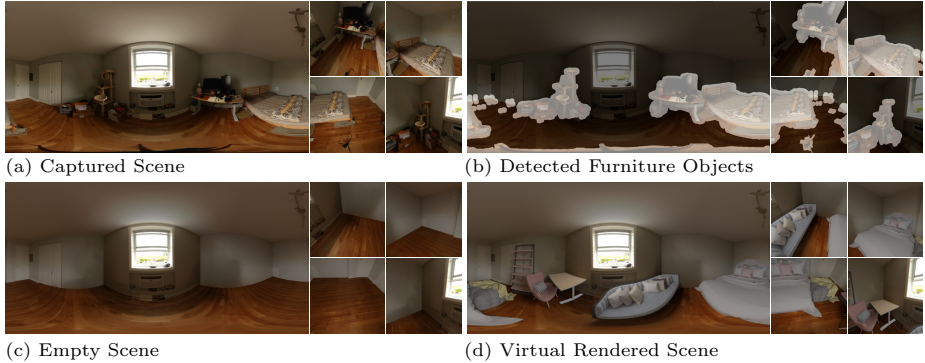
\begin{figure}
  \centering
  \begin{overpic}[width=\textwidth]{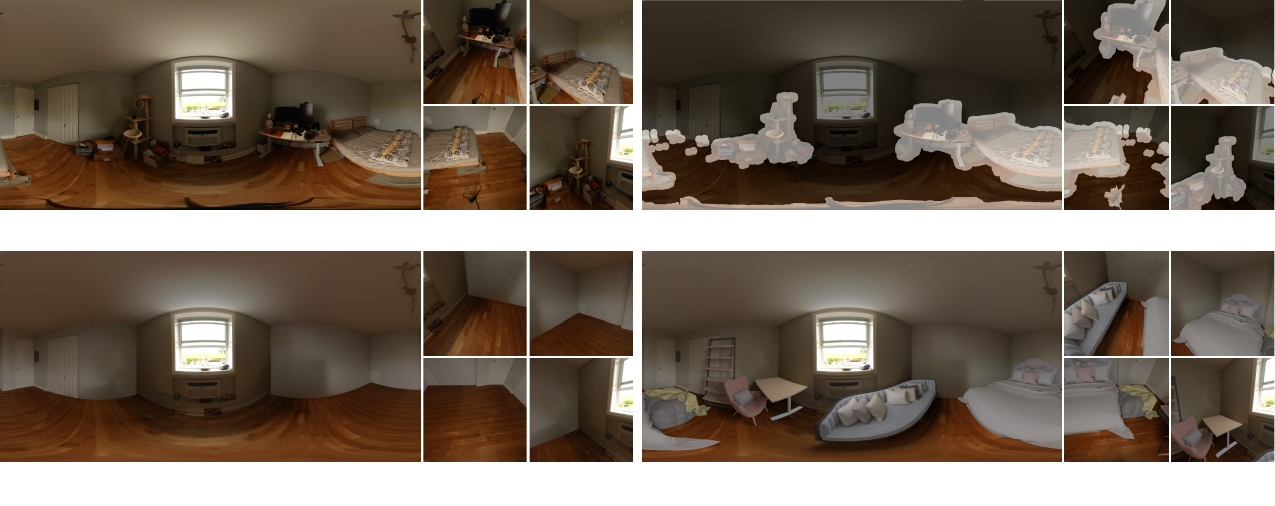}
    \put(0,21){\scriptsize \color{black}{(a) Captured Scene}}
    \put(50,21){\scriptsize \color{black}{(b) Detected Furniture Objects}}
    \put(0,1){\scriptsize \color{black}{(c) Empty Scene}}
    \put(50,1){\scriptsize \color{black}{(d) Virtual Rendered Scene}}
    
  \end{overpic}
  \caption{Illustration of Proposed Rendering Pipeline: The captured scene (a) is filled with furniture objects. Detected furniture objects (b) are removed from the scene, and an empty scene (c) is restored. (d) New furniture objects~\cite{fu20213d} are inserted and rendered with real-time outdoor illumination.}
  \label{fig_overview}
\end{figure}

\section{Introduction}
The increasing popularity of omnidirectional cameras has driven significant research interest in panoramic photography during recent years. 
A panoramic photograph provides a complete representation of the surrounding context and enables 360$^{\circ}$ rendering. The global rendering method proposed by Debevec~\cite{debevec2006image} offers a High Dynamic Range (HDR) image-based rendering model for relighting virtual objects within a realistic scene context.

Previous studies have focused on estimating 360$^{\circ}$ HDR environment map directly from Low Dynamic Range (LDR) images for scene relighting and object insertion~\cite{gardner2017learning,gardner2019deep,gkitsas2020deep}. However, these data-driven approaches often assume linear proportionality between pixel values and scene radiance without considering photometric calibration. The actual brightness of a scene, measured in luminance ($cd/m^{2}$), accurately reflects the light properties in the real world. Bolduc et al.~\cite{bolduc2023beyond} recently conducted a study that calibrated an existing panoramic HDR dataset with approximate scene luminance levels. In our work, we take this a step further by calibrating the captured HDR panoramas using absolute luminance value (in SI units) measured in each scene. This calibration ensures that our HDR images accurately represent realistic spatially varying lighting conditions, distinguishing them from existing indoor panorama datasets~\cite{xiao2012recognizing,zhang2014panocontext,cruz2021zillow}.

Panoramic images introduce unique challenges for 2D scene understanding tasks, due to the distortion caused by equirectangular projection. When dealing with scenes that contain furniture objects, the complexities of 3D scene reconstruction are further amplified. Existing image segmentation methods are primarily prepared for understanding 2D perspective images~\cite{zhou2018semantic}, limiting their applicability in panoramic images. Recent studies on indoor furniture inpainting focus on furniture removal from 2D perspective images~\cite{suvorov2022resolution,kulshreshtha2022layout}. Directly applying these inpainting techniques to furnished panoramas can result in geometric inconsistencies within indoor surfaces. Therefore, our research focuses on furniture removal tasks within panorama images and provides a restored empty room for scene editing.  

\begin{figure}[t]
  \centering
  \begin{overpic}[width=\textwidth]{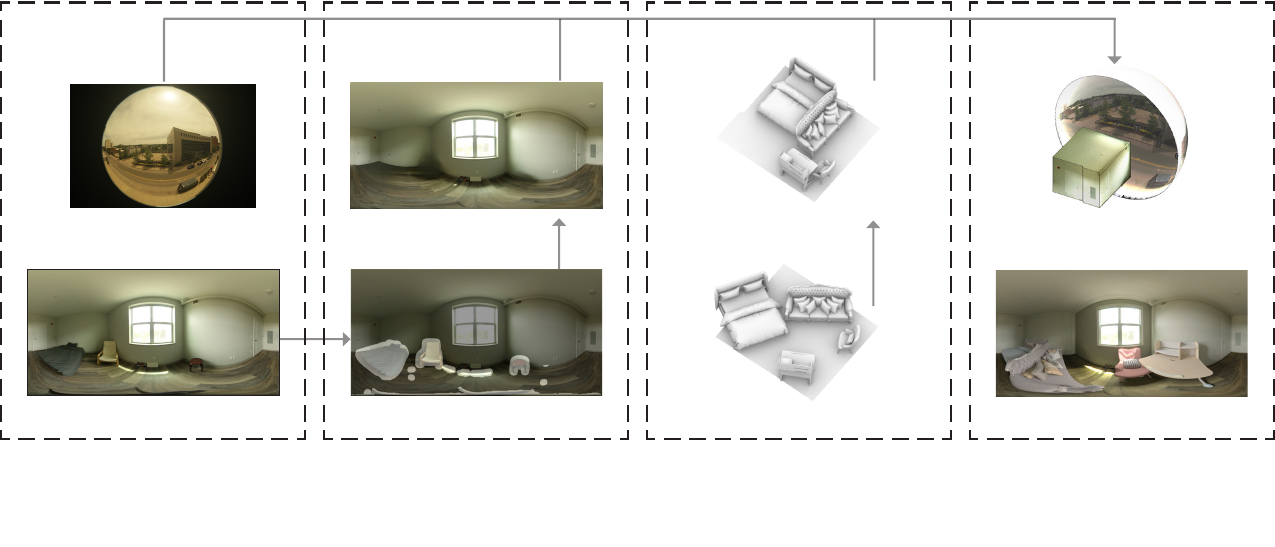}
    \put(1,25){\scriptsize \color{black}{Outdoor Photograph}}
    \put(1,10){\scriptsize \color{black}{Indoor Panorama}}
    \put(3,5){\scriptsize \color{black} \textbf{Indoor-Outdoor}}
    \put(2,3){\scriptsize \color{black} \textbf{HDR Calibration}}
    \put(7,0.5){\scriptsize \color{black} \textbf{(Sec. 3)}}
    \put(27,25){\scriptsize \color{black}{Empty Scene}}
    \put(27,10){\scriptsize \color{black}{Object Mask}}
    \put(26,5){\scriptsize \color{black} \textbf{Furniture Detection}}
    \put(29,3){\scriptsize \color{black} \textbf{and Removal}}
    \put(32,0.5){\scriptsize \color{black} \textbf{(Sec. 4)}}
    \put(52,25){\scriptsize \color{black}{Floor Layout}}
    \put(52,10){\scriptsize \color{black}{Furniture Objects}}
    \put(57,5){\scriptsize \color{black} \textbf{Automatic}}
    \put(56,3){\scriptsize \color{black} \textbf{Floor Layout}}
    \put(59,0.5){\scriptsize \color{black} \textbf{(Sec. 5)}}
    \put(78,25){\scriptsize \color{black}{Rendering Model}}
    \put(78,10){\scriptsize \color{black}{Rendered Scene}}
    \put(84,5){\scriptsize \color{black} \textbf{Indoor}}
    \put(79,3){\scriptsize \color{black} \textbf{Virtual Staging}}
    \put(84,0.5){\scriptsize \color{black} \textbf{(Sec. 6)}}
  \end{overpic}
  \caption{Our rendering pipeline consists of four modules: Indoor-Outdoor HDR Calibration (Sec. 3) calibrates the captured indoor and outdoor HDR photographs with measured absolute luminance values. Furniture Detection and Removal (Sec. 4) identifies and removes the target furniture objects from the scene. Automatic Floor Layout (Sec. 5) allows the automatic placement of multiple furniture objects. Indoor Virtual Staging (Sec. 6) achieves high-quality indoor virtual staging for furnished and empty scenes.}
  \label{fig_workflow_overview}
\end{figure}
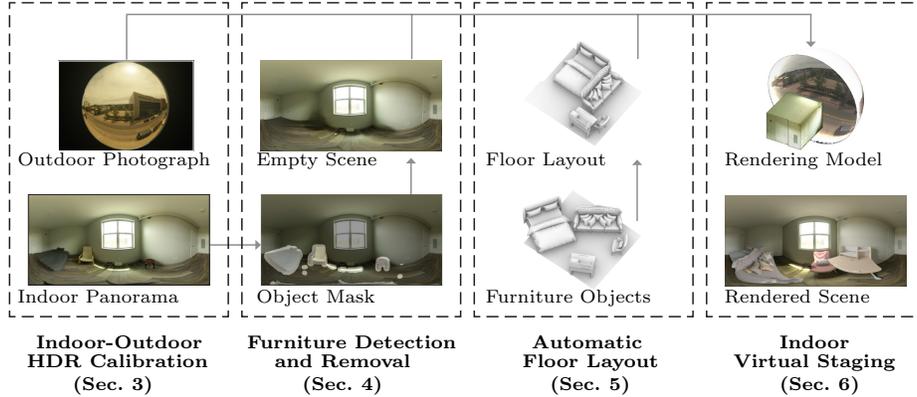

Indoor global illumination is influenced by various factors, including scene geometry, material properties, and real-time outdoor illumination. In this work, we take an existing indoor panorama and an outdoor photograph as inputs and render photo-realistic renderings featuring a new indoor furniture layout. Our rendering pipeline allows the reconstruction of global illumination between the scene and the newly inserted furniture objects (Fig~\ref{fig_workflow_overview}). In summary, this work presents the first demonstration of furniture removal, furniture insertion, and panoramic rendering for real-world indoor scenes (Fig~\ref{fig_overview}). To achieve this, our work makes the following technical contributions:
\begin{itemize}
  \item[] (1). An approach for calibrating indoor-outdoor HDR photographs and the creation of a new calibrated HDR (Cali-HDR) dataset comprising 137 scenes.
  \item[] (2). An image inpainting method that detects and removes furniture objects from a panorama.
  \item[] (3). A rule-based layout design for positioning multiple furniture objects on the floor based on spatial parameters.
\end{itemize}

\section{Related Work} 
\subsubsection{HDR and Photometric Calibration} 
The dynamic range of radiances in a real-world scene spans from $10^{-3}$ $cd/m^{2}$ (starlight) to $10^{5}$ $cd/m^{2}$ (sunlight)~\cite{reinhard2010high}. In the context of a 2D perspective image, some studies have focused on predicting panoramic HDR environment maps~\cite{gardner2017learning}, lighting representation~\cite{gardner2019deep}, and estimating HDR panoramas from LDR images~\cite{gkitsas2020deep}. Considering that HDR images reflect the relative luminance values from the real world, absolute luminance measurement is required for on-site HDR photography to recover scene radiance~\cite{debevec2008recovering}. To display the absolute luminance value, the captured HDR image requires photometric calibration, which is a means of radiometric self-calibration~\cite{mitsunaga1999radiometric}. Reference planes, such as matte color checkers or gray cards, should be positioned within the scene for luminance measurement~\cite{moeck2007accuracy}. 

\subsubsection{Indoor Light Estimation} 
Previous studies on indoor lighting estimation have explored indoor lighting editing~\cite{li2022physically}, material property estimation~\cite{yeh2022photoscene}, and the recovery of spatially-varying lighting~\cite{li2020inverse,garon2019fast,srinivasan2020lighthouse} from a 2D image. Following the global rendering method~\cite{debevec2008rendering}, some studies aim to estimate a 360$^{\circ}$ indoor HDR environment map from a 2D image and subsequently render the virtual objects~\cite{legendre2019deeplight,gardner2017learning}. 
User inputs, such as annotating indoor planes and light sources, have also been utilized to assist scene relighting and object insertion~\cite{karsch2011rendering}. Zhi et al. decompose the light effects in the empty panoramas~\cite{zhi2022semantically}. While previous studies have extensively focused on global light estimation and 3D object insertion, there is limited research on panoramic global rendering under real-time outdoor illumination.

\subsubsection{Panoramic Furniture Removal} 
The conventional image inpainting method assumes a nearly planar background around the target object, making it unsuitable for indoor scenes with complex 3D room structures. For the case of indoor scenes, even state-of-the-art inpainting models, such as LaMa~\cite{suvorov2022resolution}, cannot recognize the global structure, including the boundaries of walls, ceilings, and floors. Several approaches have been attempted to address this challenge: (1) utilizing lighting and geometry constraints~\cite{zhang2021no}, (2) using planar surfaces to approximate contextual geometry~\cite{kawai2015diminished,huang2014image,kulshreshtha2022layout}, and (3) estimating an empty 3D room geometry from furnished scenes~\cite{zhang2016emptying}. These studies have primarily focused on furniture detection and inpainting tasks for 2D perspective images. 
Panoramic scene understanding includes object detection~\cite{guerrero2020s} and spherical semantic segmentation~\cite{zhang2019orientation}.
Although the recent studies~\cite{gkitsas2021panodr,gkitsas2021towards} have started furniture removal tasks in panoramas, it is primarily centered around virtually rendered scenes rather than real-world scenes. 

\subsubsection{3D Layout Estimation} 
Estimating a 3D room layout from a single image is a common task for indoor scene understanding. While indoor panorama can be converted into a cubic map~\cite{cheng2018cube,BiFuse20}, the actual 3D layout is oversimplified. Building on this cube map approach, other studies~\cite{BiFuse20,wang2018self} focus on panorama depth estimation using 3D point clouds. Moreover, under the Manhattan world assumption~\cite{coughlan1999manhattan}, a 360$^{\circ}$ room layout with separated planar surfaces can be segmented from a single panorama~\cite{wang2021led2,zou2018layoutnet,zhang2014panocontext,yang2019dula}. Moving beyond 3D room layout, detailed scene and furniture geometry can be reconstructed from 2D perspective images~\cite{huang2018holistic,izadinia2017im2cad,nie2020total3dunderstanding}. Additionally, when provided with a 2D floor plan image, indoor space semantics and topology representations can be generated to create a 3D model~\cite{yang2022automated} and recognize elements in floor layouts~\cite{zeng2019deep}. An accurate room geometry allows new furniture objects to be inserted precisely into the existing scene. 

\section{Indoor-Outdoor HDR Calibration}
\label{sec:cali_hdr}
\subsubsection{Indoor HDR  Calibration} For indoor scenes, a Ricoh Theta Z1 camera was positioned in the room to capture panoramic HDR photographs. The camera settings were configured as follows: White Balance (Daylight 6500), ISO (100), Aperture (F/5.6), Image Size (6720 x 3360), and Shutter Speed (4, 1, 1/4, 1/15, 1/60, 1/250, 1/1000, 1/4000, 1/8000). To ensure consistency and avoid motion blur during photography, the camera was fixed on a tripod at a height of 1.6$m$. We placed a Konica Minolta LS-160 luminance meter next to the camera to measure the target luminance on a white matte board. Each HDR photograph needs per-pixel calibration to accurately display luminance values for the scene. The measured absolute luminance value at the selected point is recorded in SI unit ($cd/m^{2}$). The measured luminance value and displayed luminance value from the original HDR image are used for calculating the calibration factor ($k_1$). According to the study by Inanici~\cite{inanici2006evaluation}, given $R$, $G$, and $B$ values in the captured indoor HDR image, indoor scene luminance ($L_i$) is expressed as: 
\begin{equation}
  L_i =  k_1 \cdot (0.2127 \cdot R + 0.7151 \cdot G + 0.0722 \cdot B) (cd/m^{2}) 
\end{equation}  

\subsubsection{Outdoor HDR  Calibration} To capture outdoor scenes, a Canon EF 8-15mm f/4L fisheye lens was installed on Canon EOS 5D Mark II Full Frame DSLR Camera, and a 3.0 Neutral Density (ND) filter was utilized for capturing direct sunlight with HDR technique~\cite{stumpfel2006direct}. The camera settings were configured as follows: White Balance (Daylight 6500), ISO (200), Aperture (F/16), Image Size (5616 x 3744), and Shutter Speed (4, 1, 1/4, 1/15, 1/60, 1/250, 1/1000, 1/4000, 1/8000). Due to the diverse outdoor contexts, it is impractical to place a target plane to measure target luminance values. Each camera has its own fixed camera response curve to merge multiple images with varying exposures into one single HDR image. Rather than performing a separate calibration process for outdoor HDR, our objective is to determine a fixed calibration factor between two distinct cameras and calibrate the outdoor HDR images with indoor luminance measurement.
\begin{figure}
  \centering
  \begin{overpic}[width=\textwidth]{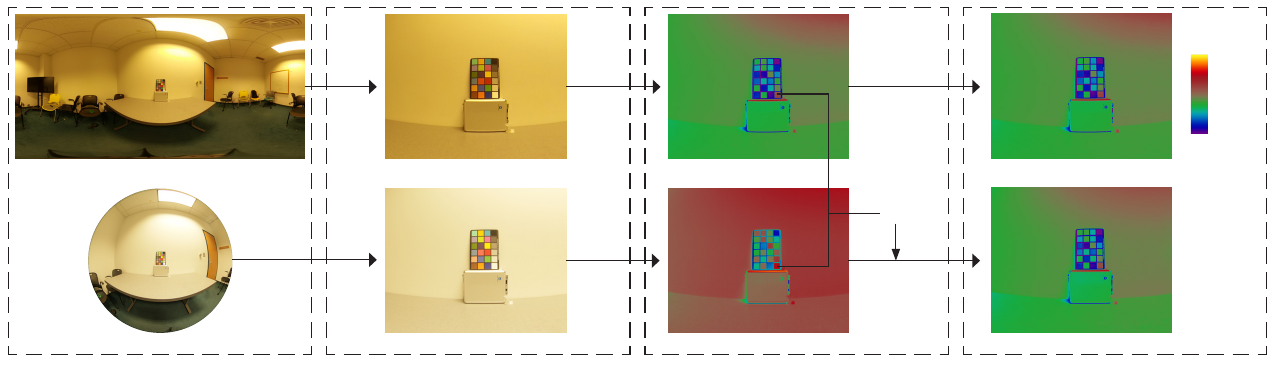}
    \put(10,0){\small \color{black}{(a)}}
    \put(2,17){\tiny \color{black}{Ricoh Theta Z1}}  
    \put(2,3.5){\tiny \color{black}{Canon Fisheye Lens}}  
    \put(35,0){\small \color{black}{(b)}}
    \put(60,0){\small \color{black}{(c)}}
    \put(85,0){\small \color{black}{(d)}}
    \put(70,14){\small \color{black}{$k_2$}}
    \put(93,29){\tiny \color{black}{5,000}}    
    \put(93,27.5){\tiny \color{black}{$cd/m^{2}$}}   
    \put(93,19){\tiny \color{black}{50}}    
    \put(93,17.5){\tiny \color{black}{$cd/m^{2}$}} 
  \end{overpic}
  \caption{Calibration Process: (a) Photographs from two cameras. (b) Cropped target regions. (c) Original luminance maps. (d) Luminance maps after HDR image captured by fisheye lens is scaled with $k_2$.}
  \label{fig_fisheye2hdr}
\end{figure}
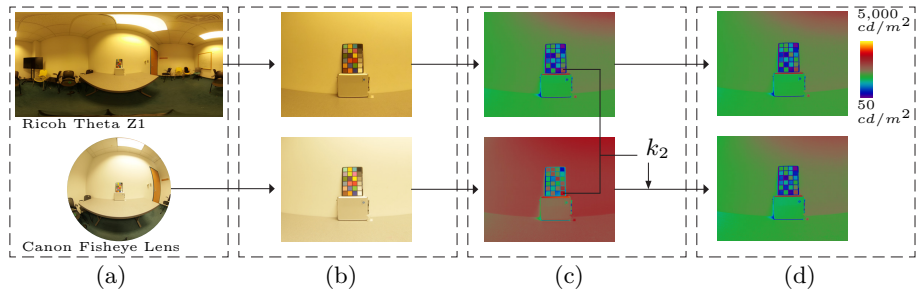

As shown in Fig.~\ref{fig_fisheye2hdr}, we positioned two cameras in an enclosed room under consistent electrical lighting. Following the camera settings of indoor and outdoor HDR photography (Sec.~\ref{sec:cali_hdr}), we captured the target checkboard from two cameras, respectively. Then, 2D perspective images displaying the same target were cropped from the original images. After merging the two sets of images into HDR photographs, we calculated the difference ratio ($k_2$) between the target pixel region (white patch) on the HDR photographs obtained from the two cameras. Ultimately, the HDR image captured by Canon EOS 5D Camera was linearly calibrated with the computed constant value ($k_2$), and the HDR photographs from the two cameras were calibrated to display the same luminance range. $k_2$ is a fixed constant when the two camera settings stay the same. Given $R$, $G$, and $B$ values in the captured outdoor HDR image, outdoor scene luminance ($L_o$) is expressed as:
\begin{equation}
  L_o =  k_1 \cdot k_2 \cdot (0.2127 \cdot R + 0.7151 \cdot G + 0.0722 \cdot B) (cd/m^{2}) 
\end{equation}

where $k_1$ is the calibration factor determined by the measured luminance target value and displayed luminance value in the captured indoor HDR image, and $k_2$ is the computed constant for scaling the outdoor hemispherical image into the indoor panorama. 

After linear rescaling, the outdoor HDR photographs are processed through the following steps: (1) vignetting correction that compensates for the light loss in the periphery area caused by the fisheye lens~\cite{inanici2006evaluation}, (2) color correction for chromatic changes introduced by ND filter~\cite{stumpfel2006direct}, and (3) geometric transformation from equi-distant to hemispherical fisheye image for environment mapping~\cite{inanici2010evalution}.  

\section{Furniture Detection and Removal}
\label{sec:pano_furn_detect}
\subsubsection{Panoramic Furniture Detection} A single panorama displayed in 2D image coordinates can be transformed into a 3D spherical representation~\cite{wang2021led2,araujo2018drawing}, and this process can also be inverted. Building on this concept, our objective is to convert a panorama into a list of 2D images for scene segmentation. Subsequently, we aim to reconstruct the panorama where target furniture objects are highlighted. The selected region on the input panorama $I_p$ is geometrically cropped and transformed into a 2D perspective image, within longitude angle ($\theta$) and latitude angle ($\phi$). $\theta \in (-\pi, +\pi)$  and $\phi \in (-0.5\pi, +0.5\pi)$. With the fixed field of view ($FOV$) and the image dimension of height ($h$) by width ($w$), we obtain 2D perspective image set $\mathrm{I} = \{I_1, I_2, I_3, \ldots, I_i\}$, and the process of equirectangular-to-perspective can be expressed as mapping function $S$:  

\begin{equation} 
  I_i = S(I_p; \, FOV,\,\theta,\, \phi,\, h,\, w)
\end{equation}

\begin{figure}
  \centering
  \begin{overpic}[width=\textwidth]{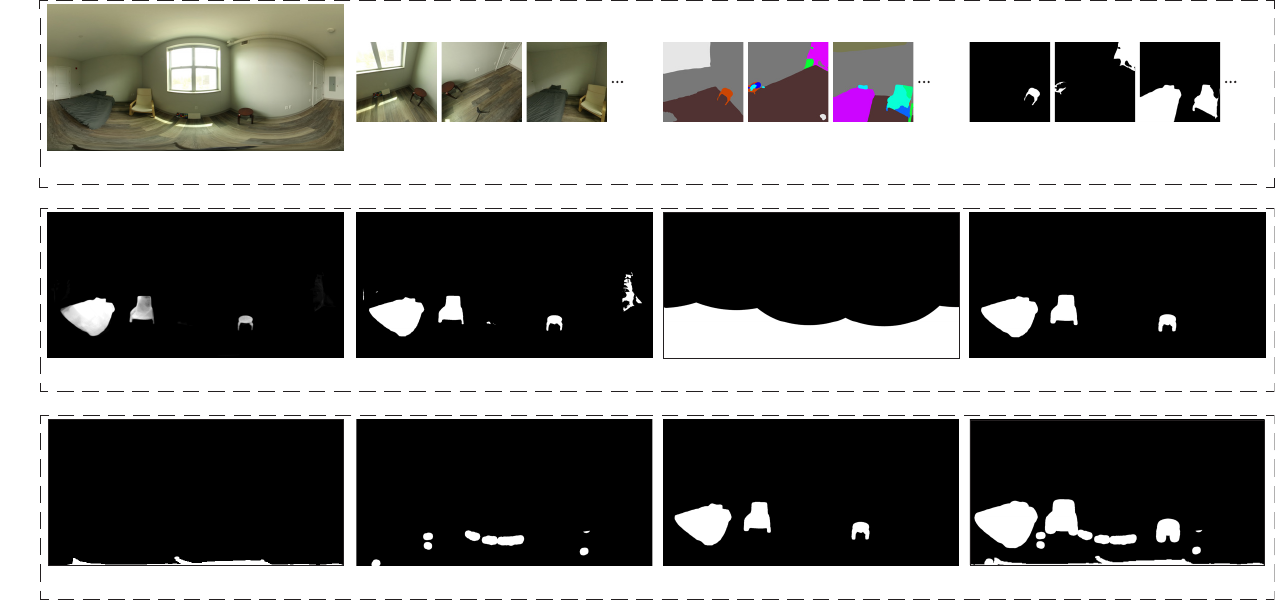}
    \put(-1,40){\small \color{black}{(a)}}
    \put(4,34){\scriptsize \color{black}{Panorama}}
    \put(28,34){\scriptsize \color{black}{2D Perspective}}
    \put(52,34){\scriptsize \color{black}{Segmentation}}
    \put(76,34){\scriptsize \color{black}{Binarization}}
    \put(-1,23){\small \color{black}{(b)}}
    \put(4,17){\scriptsize \color{black}{Stitched Panorama}}
    \put(28,17){\scriptsize \color{black}{Filtered Panorama}}
    \put(52,17){\scriptsize \color{black}{Floor Boundary}}
    \put(76,17){\scriptsize \color{black}{Detected Furniture}}
    \put(-1,7){\small \color{black}{(c)}}
    \put(4,1){\scriptsize \color{black}{Tripod Location}}
    \put(28,1){\scriptsize \color{black}{Sunlight Region}}
    \put(52,1){\scriptsize \color{black}{Furniture Areas}}
    \put(76,1){\scriptsize \color{black}{Target Mask}}
  \end{overpic}
  \caption{Panoramic Scene Segmentation: (a) A single panorama is segmented into a set of 2D perspective images, and target furniture objects are detected. (b) The stitched panorama is processed to display furniture contours, and the rendered floor boundary is utilized to filter out solid contours that are not attached to the floor area. (c) Estimated tripod location~\cite{zhi2022semantically}, direct sunlight region, and the detected furniture areas are combined as the target mask.}
  \label{fig_pano_furn_det}
\end{figure}

After scene segmentation for 2D perspective images, a set of processed images $\mathrm{I}' = \{I_1', I_2', I_3', \ldots, I_i'\}$ is stitched back to reconstruct a new panorama according to annotated $\theta$ and $\phi$. The invertible mapping process enables image transformation between equirectangular and 2D perspective representations. As shown in Fig.~\ref{fig_pano_furn_det}, one single panorama is segmented into a set of 2D perspective images and segmented per color scheme in semantic segmentation classes~\cite{zhou2018semantic}. Given a furnished panorama (Fig.~\ref{fig_pano_furn_det}(a)), a 3D layout is estimated with separated planer surfaces of the ceiling, wall, and floor textures. The rendering model generates an indoor mask to distinguish the floor and other interior surfaces, and the result highlights the furniture object placed on the floor (Fig.~\ref{fig_pano_furn_det}(b)). 

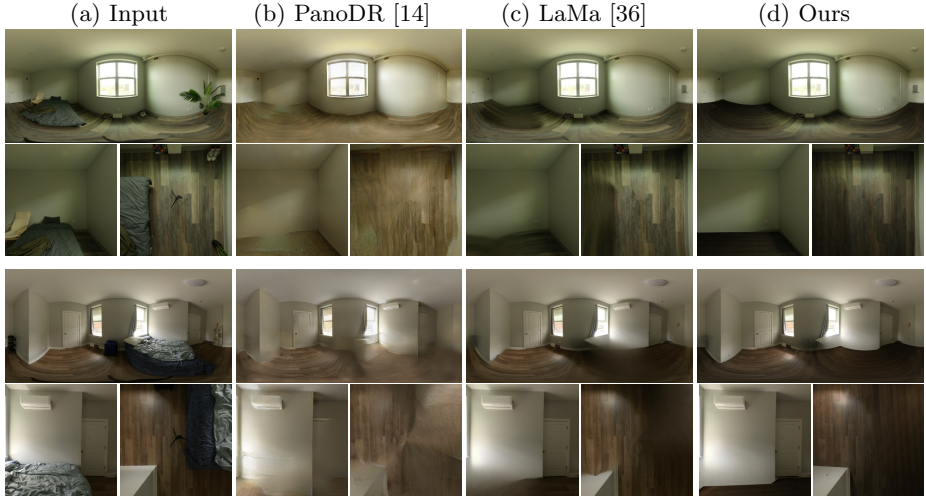
\begin{figure}
  \centering
  \begin{overpic}[width=\textwidth]{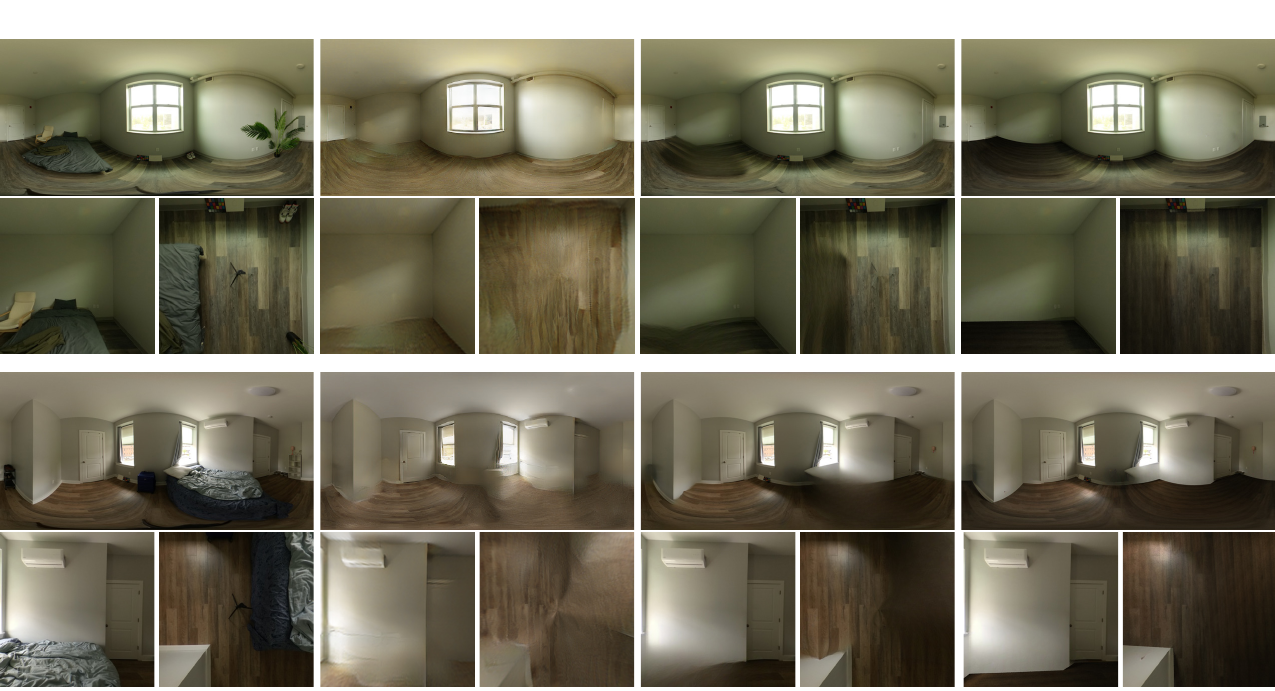}
    \put(7,52){\small \color{black}{(a) Input}}
    \put(27,52){\small \color{black}{(b) PanoDR~\cite{gkitsas2021panodr}}}
    \put(54,52){\small \color{black}{(c) LaMa~\cite{suvorov2022resolution}}}
    \put(82,52){\small \color{black}{(d) Ours}}
  \end{overpic}
  \caption{Comparison of Image Inpainting Methods: The target mask (from Fig.~\ref{fig_pano_furn_det}(c)) is paired with input panorama (a) to remove the target region using PanoDR~\cite{gkitsas2021panodr}(b), LaMa~\cite{suvorov2022resolution}(c), and our method (d), respectively.}
  \label{fig_fur_rem_compare}
\end{figure}

\subsubsection{Furniture Removal} For furnished panoramas, we first estimate the 3D room geometry~\cite{wang2021led2} and utilize the indoor planar information in the panoramas to guide the inpainting process. As shown in Fig.~\ref{fig_fur_rem_compare}, our method allows for image inpainting on the original furnished panoramas with surrounding context, while utilizing the floor boundary as a guiding reference to preserve clear indoor boundaries. One challenge in inpainting the floor texture is when the masked region is distant from nearby pixels, leading to blurring and noise. Unlike walls and ceilings, the floor texture often exhibits a strong pattern with various textures. Thus, we address this issue by treating the floor texture in the indoor scenes as a Near-Periodic Pattern (NPP). 
Compared to LaMa~\cite{suvorov2022resolution}, which is trained on existing 2D image datasets, the NPP model developed by Chen et al.~\cite {chen2022learning} learns the masked region from the provided image. This results in outputs that are optimized based on the content of the input image itself. As demonstrated in Fig.~\ref{fig_fur_rem_compare}, our approach, combined with the LaMa~\cite{suvorov2022resolution} and NPP models~\cite{chen2022learning}, effectively recovers the scene context around the detected furniture area. The restored indoor textures, including ceiling, wall and floors, will be incorporated into the 3D rendering model. 

\section{Automatic Floor Layout}
\label{sec:flr_design}
The rendering model comprises 3D room geometry, allowing precise placement of multiple furniture objects with different orientations and positions. The floor layout follows a series of spatial parameters and rules for furniture arrangements. We segment the floor mesh from the panorama, and the orientation of each object is determined based on whether it faces the window or indoor walls. For the translation distance, we normalize the distance between the object's dimension and the floor boundary to a range between 0 and 1. This normalization allows the object to be precisely positioned along the wall and window side. Different spatial parameters and orientation combinations can express alternative floor layouts. The rule-based method adapts to various layout rules by recognizing different floor boundaries and placing target objects accordingly within different indoor scenes. 

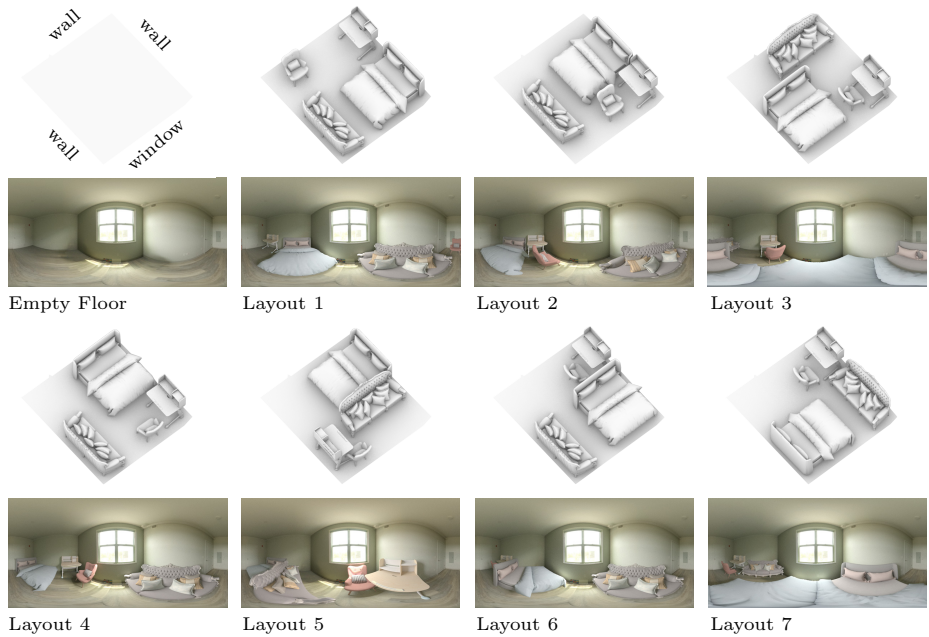
\begin{figure}
  \centering
  \begin{overpic}[width=\textwidth]{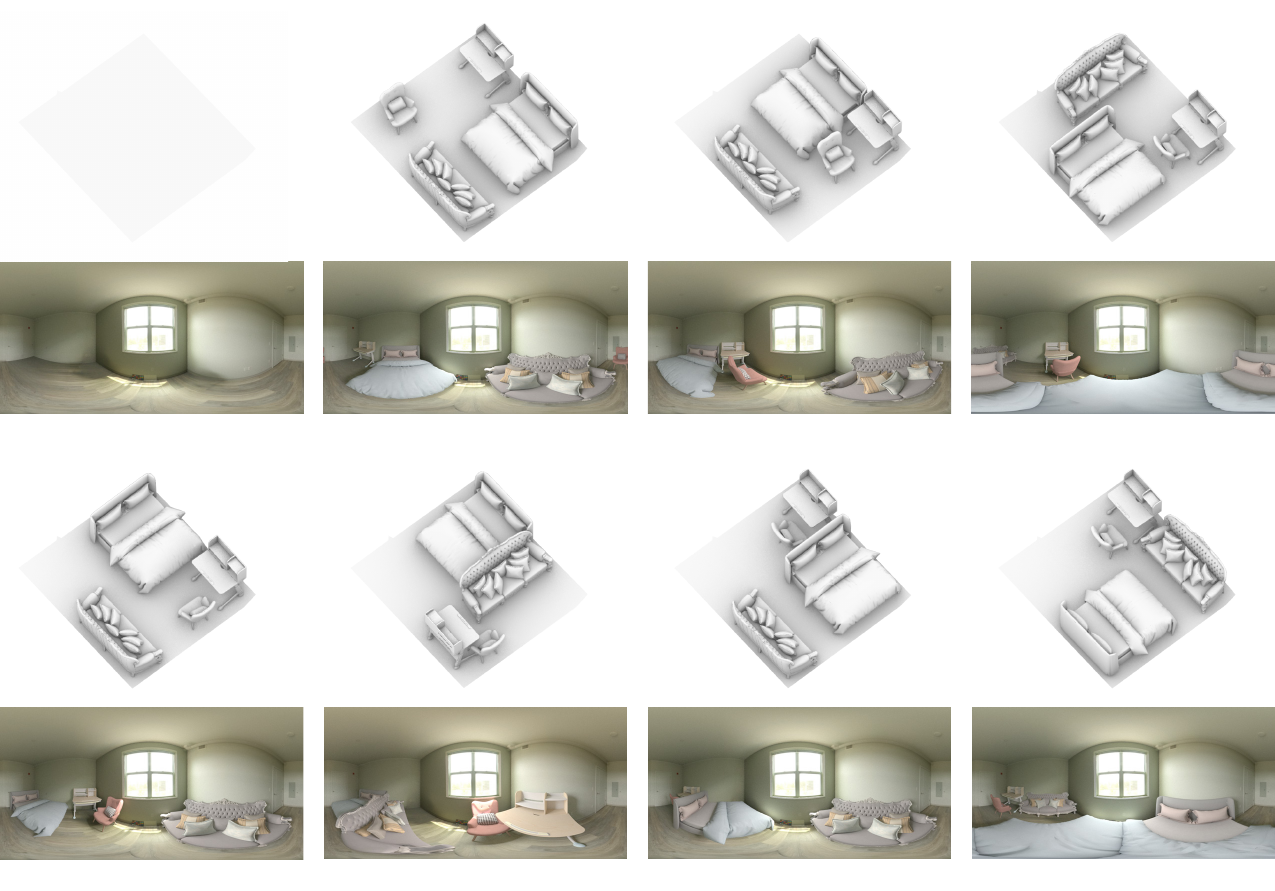}
    \put(6,53){\scriptsize{\makebox(0,0){\rotatebox{-48}{wall}}}}
    \put(16,53){\scriptsize{\makebox(0,0){\rotatebox{41}{window}}}}
    \put(16,65){\scriptsize{\makebox(0,0){\rotatebox{-48}{wall}}}}
    \put(6,66){\scriptsize{\makebox(0,0){\rotatebox{41}{wall}}}}
    \put(0,35){\scriptsize \color{black}{Empty Floor}}
    \put(25.5,35){\scriptsize \color{black}{Layout 1}}
    \put(51,35){\scriptsize \color{black}{Layout 2}}
    \put(76.5,35){\scriptsize \color{black}{Layout 3}}
    \put(0,0){\scriptsize \color{black}{Layout 4}}
    \put(25.5,0){\scriptsize \color{black}{Layout 5}}
    \put(51,0){\scriptsize \color{black}{Layout 6}}
    \put(76.5,0){\scriptsize \color{black}{Layout 7}}  
  \end{overpic}
   \caption{Furniture Layout Alternatives: Given an empty floor mesh, multiple furniture objects are placed on the floor with predefined positions and orientations.}
  \label{fig_auto_layout}
\end{figure}

Within the 3D coordinate system, the segmented floor mesh and furniture objects are positioned on the $xy$ plane (Fig~\ref{fig_auto_layout}). Each furniture object can be represented as a set of point clouds. The task of floor layout design is subject to the constraint of the floor boundary. Each furniture object rotates around the $z$ axis by an angle $\theta$ to align with the target floor edge and translates itself to the designated position, denoted by the distances $t_x$ and $t_y$. We transform the 3D point set $\mathbf{x_i}$ to its corresponding transformed point $\mathbf{x_i}'$ in the $xy$ plane, by applying the rotation matrix and the translation matrix: $\mathbf{x_i}' = R_z (\theta) \mathbf{x_i} + t$, where $t = \begin{bmatrix}
                            t_x &
                            t_y &
                            0 &
                            \end{bmatrix}^T $.

\section{Indoor Virtual Staging}
\begin{figure}
  \centering
  \begin{overpic}[width=\textwidth]{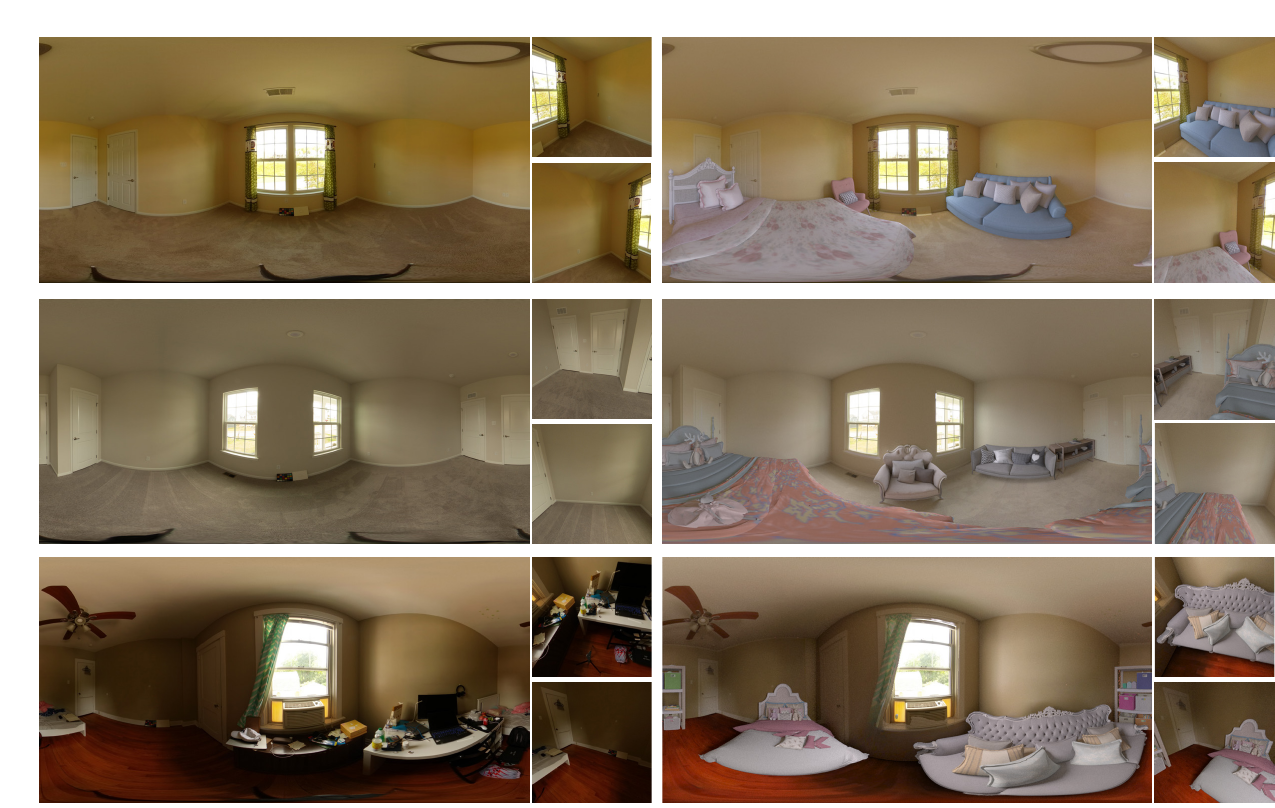}
    \put(1,51){\scriptsize{\makebox(0,0){\rotatebox{90}{Room 1}}}}
    \put(1,31){\scriptsize{\makebox(0,0){\rotatebox{90}{Room 2}}}}
    \put(1,11){\scriptsize{\makebox(0,0){\rotatebox{90}{Room 3}}}}
    \put(13,61){\scriptsize \color{black}{Captured Scenes}}
    \put(60,61){\scriptsize \color{black}{Virtual Rendered Scenes}}
  \end{overpic}
  \caption{Photo Gallery of Scene Editing: (left) The captured scenes include empty and furnished rooms. (right) The scenes are virtually rendered with new furniture objects~\cite{fu20213d}.}
  \label{fig_scene_redesign}
\end{figure}

We tested our methodology in various real-world scenes and refurnished the existing scenes with virtual furniture objects (Fig~\ref{fig_scene_redesign}). The new virtual scenes are rendered under real-time outdoor illumination. Compared to previous scene relighting and object insertion approaches~\cite{zhi2022semantically,li2020inverse,gardner2017learning}, our proposed rendering method integrates complete 3D scene geometry (including both room geometry and furniture objects), outdoor environment map, and material textures. This rendering setup allows the new furniture objects to be virtually rendered within the scene. 

\begin{figure}
  \centering
  \begin{overpic}[width=\textwidth]{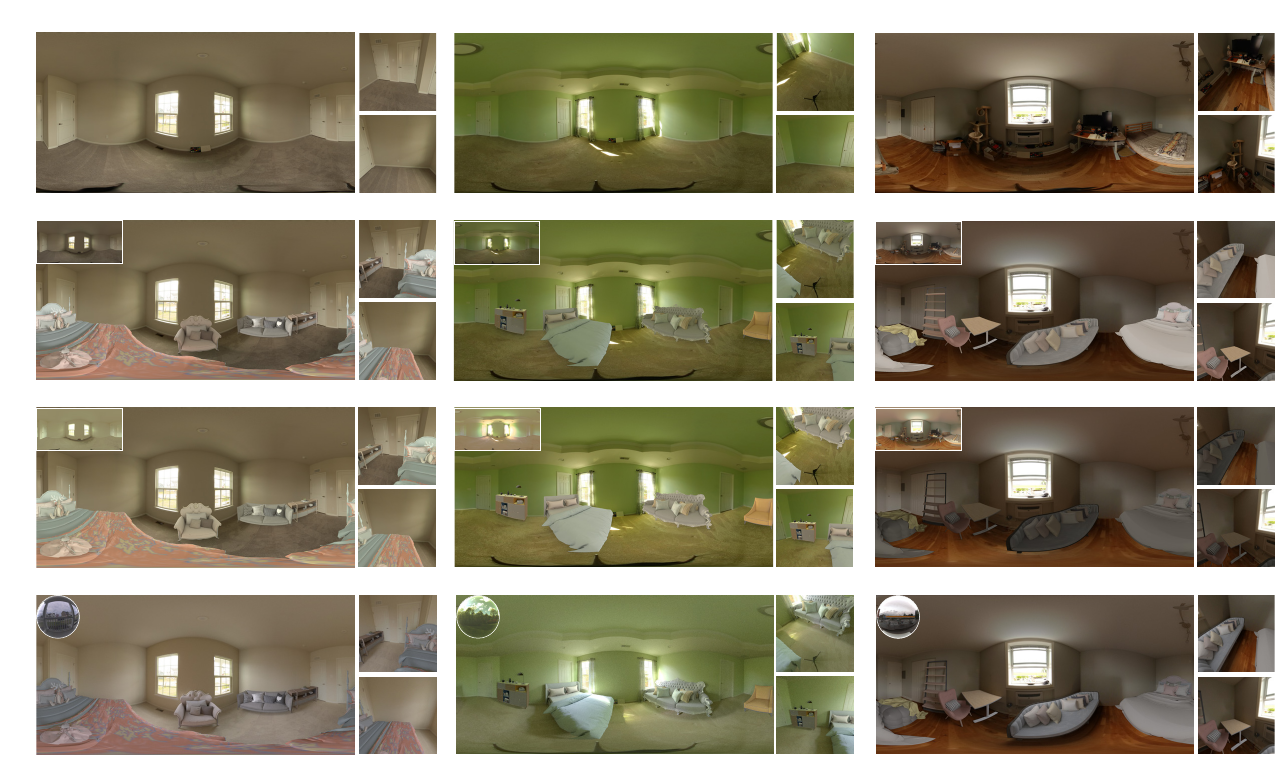}
      \put(11,57.5){\scriptsize \color{black}{Room 1}}
      \put(44,57.5){\scriptsize \color{black}{Room 2}}
      \put(77,57.5){\scriptsize \color{black}{Room 3}}
      \put(-1,50){\scriptsize \color{black}{(a)}}
      \put(-1,35){\scriptsize \color{black}{(b)}}
      \put(-1,20){\scriptsize \color{black}{(c)}}
      \put(-1,5){\scriptsize \color{black}{(d)}}
  \end{overpic}
  \caption{Comparison of Different Rendering Methods: LDR panoramas (a) are used to generate HDR panoramas (b) by Liu et al.'s method~\cite{liu2020single} and render furniture objects. Furniture objects are rendered by our calibrated indoor HDR panoramas (c). Virtual scenes are rendered by our method (d).}
  \label{fig_global_ill}
\end{figure}
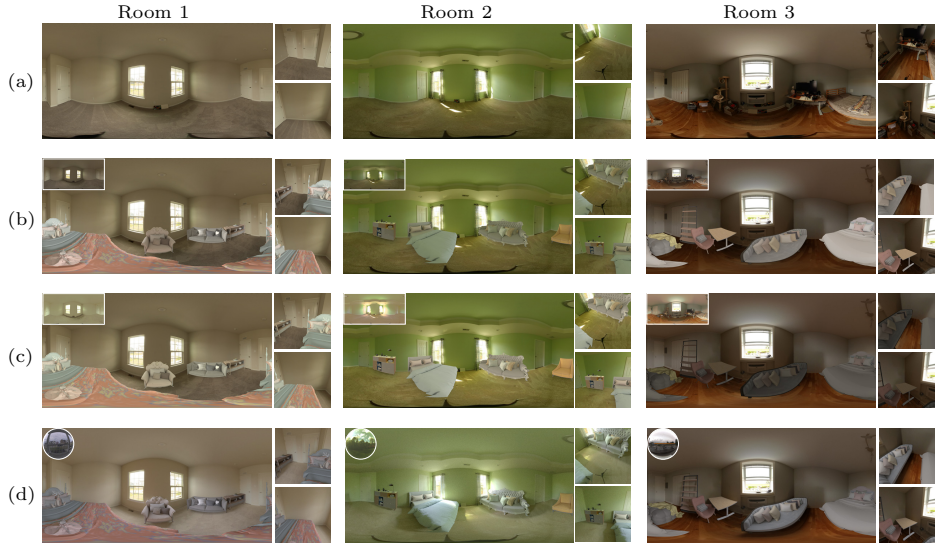

By using the real-time outdoor HDR image as the light source, we achieve realistic global illumination within the indoor space and reconstruct the indoor scenes with corresponding outdoor lighting conditions (Fig~\ref{fig_global_ill}). The proposed rendering approach not only accurately renders the virtual furniture objects but also reconstructs the inter-reflection between the scene and newly inserted objects. It is important to note that as the scene geometry is approximated into individual planar surfaces, certain indoor details such as curtains or window frames are simplified in the rendering model. Overall, our rendering pipeline effectively generates high-quality indoor panoramas while preserving the essential characteristics of the real-world scenes.

\section{Conclusion and Limitation}
In this paper, we presented a complete rendering framework that effectively transforms an existing panorama into a new furnished scene, providing high-quality virtual panoramas for 360$^{\circ}$ virtual staging. Additionally, we introduce a parametric modeling method for placing multiple furniture objects within the scene, which improves the flexibility of floor layout design. The global rendering framework offers a robust solution for realistic virtual home staging and contributes new indoor rendering techniques.

Some limitations exist in our study. The current implementation of the automatic floor layout does not account for the presence of doors in the scene. This means that the generated floor layouts may not fully account for the locations of the doors, potentially leading to impractical furniture arrangements. Furthermore, our research was limited to a fixed view position to match the captured panorama. To expand on these findings, future work will investigate varying view positions within the indoor space and explore human's visual perception under different illumination conditions.

\section*{Acknowledgement} This work was partially supported by a gift from Zillow Group, USA. 

\bibliographystyle{splncs04}
\bibliography{references}

\begin{thebibliography}{10}
\providecommand{\url}[1]{\texttt{#1}}
\providecommand{\urlprefix}{URL }
\providecommand{\doi}[1]{https://doi.org/#1}

\bibitem{araujo2018drawing}
Ara{\'u}jo, A.B.: Drawing equirectangular vr panoramas with ruler, compass, and
  protractor. Journal of Science and Technology of the Arts  \textbf{10}(1),
  15--27 (2018)

\bibitem{bolduc2023beyond}
Bolduc, C., Giroux, J., H{\'e}bert, M., Demers, C., Lalonde, J.F.: Beyond the
  pixel: a photometrically calibrated hdr dataset for luminance and color
  temperature prediction. arXiv preprint arXiv:2304.12372  (2023)

\bibitem{chen2022learning}
Chen, B., Zhi, T., Hebert, M., Narasimhan, S.G.: Learning continuous implicit
  representation for near-periodic patterns. In: Computer Vision--ECCV 2022:
  17th European Conference, Tel Aviv, Israel, October 23--27, 2022,
  Proceedings, Part XV. pp. 529--546. Springer (2022)

\bibitem{cheng2018cube}
Cheng, H.T., Chao, C.H., Dong, J.D., Wen, H.K., Liu, T.L., Sun, M.: Cube
  padding for weakly-supervised saliency prediction in 360 videos. In:
  Proceedings of the IEEE Conference on Computer Vision and Pattern
  Recognition. pp. 1420--1429 (2018)

\bibitem{coughlan1999manhattan}
Coughlan, J.M., Yuille, A.L.: Manhattan world: Compass direction from a single
  image by bayesian inference. In: Proceedings of the seventh IEEE
  international conference on computer vision. vol.~2, pp. 941--947. IEEE
  (1999)

\bibitem{cruz2021zillow}
Cruz, S., Hutchcroft, W., Li, Y., Khosravan, N., Boyadzhiev, I., Kang, S.B.:
  Zillow indoor dataset: Annotated floor plans with 360deg panoramas and 3d
  room layouts. In: Proceedings of the IEEE/CVF Conference on Computer Vision
  and Pattern Recognition. pp. 2133--2143 (2021)

\bibitem{debevec2006image}
Debevec, P.: Image-based lighting. In: ACM SIGGRAPH 2006 Courses, pp. 4--es
  (2006)

\bibitem{debevec2008rendering}
Debevec, P.: Rendering synthetic objects into real scenes: Bridging traditional
  and image-based graphics with global illumination and high dynamic range
  photography. In: siggraph 2008 classes, pp. 1--10. ACM (2008)

\bibitem{debevec2008recovering}
Debevec, P.E., Malik, J.: Recovering high dynamic range radiance maps from
  photographs. In: SIGGRAPH 2008 classes, pp. 1--10. ACM (2008)

\bibitem{fu20213d}
Fu, H., Jia, R., Gao, L., Gong, M., Zhao, B., Maybank, S., Tao, D.: 3d-future:
  3d furniture shape with texture. International Journal of Computer Vision pp.
  1--25 (2021)

\bibitem{gardner2019deep}
Gardner, M.A., Hold-Geoffroy, Y., Sunkavalli, K., Gagn{\'e}, C., Lalonde, J.F.:
  Deep parametric indoor lighting estimation. In: Proceedings of the IEEE/CVF
  International Conference on Computer Vision. pp. 7175--7183 (2019)

\bibitem{gardner2017learning}
Gardner, M.A., Sunkavalli, K., Yumer, E., Shen, X., Gambaretto, E., Gagn{\'e},
  C., Lalonde, J.F.: Learning to predict indoor illumination from a single
  image. arXiv preprint arXiv:1704.00090  (2017)

\bibitem{garon2019fast}
Garon, M., Sunkavalli, K., Hadap, S., Carr, N., Lalonde, J.F.: Fast
  spatially-varying indoor lighting estimation. In: Proceedings of the IEEE/CVF
  Conference on Computer Vision and Pattern Recognition. pp. 6908--6917 (2019)

\bibitem{gkitsas2021panodr}
Gkitsas, V., Sterzentsenko, V., Zioulis, N., Albanis, G., Zarpalas, D.: Panodr:
  Spherical panorama diminished reality for indoor scenes. In: Proceedings of
  the IEEE/CVF Conference on Computer Vision and Pattern Recognition. pp.
  3716--3726 (2021)

\bibitem{gkitsas2020deep}
Gkitsas, V., Zioulis, N., Alvarez, F., Zarpalas, D., Daras, P.: Deep lighting
  environment map estimation from spherical panoramas. In: Proceedings of the
  IEEE/CVF Conference on Computer Vision and Pattern Recognition Workshops. pp.
  640--641 (2020)

\bibitem{gkitsas2021towards}
Gkitsas, V., Zioulis, N., Sterzentsenko, V., Doumanoglou, A., Zarpalas, D.:
  Towards full-to-empty room generation with structure-aware feature encoding
  and soft semantic region-adaptive normalization. arXiv preprint
  arXiv:2112.05396  (2021)

\bibitem{guerrero2020s}
Guerrero-Viu, J., Fernandez-Labrador, C., Demonceaux, C., Guerrero, J.J.:
  What’s in my room? object recognition on indoor panoramic images. In: 2020
  IEEE International Conference on Robotics and Automation (ICRA). pp.
  567--573. IEEE (2020)

\bibitem{huang2014image}
Huang, J.B., Kang, S.B., Ahuja, N., Kopf, J.: Image completion using planar
  structure guidance. ACM Transactions on graphics (TOG)  \textbf{33}(4),
  1--10 (2014)

\bibitem{huang2018holistic}
Huang, S., Qi, S., Zhu, Y., Xiao, Y., Xu, Y., Zhu, S.C.: Holistic 3d scene
  parsing and reconstruction from a single rgb image. In: Proceedings of the
  European conference on computer vision (ECCV). pp. 187--203 (2018)

\bibitem{inanici2010evalution}
Inanici, M.: Evalution of high dynamic range image-based sky models in lighting
  simulation. Leukos  \textbf{7}(2),  69--84 (2010)

\bibitem{inanici2006evaluation}
Inanici, M.N.: Evaluation of high dynamic range photography as a luminance data
  acquisition system. Lighting Research \& Technology  \textbf{38}(2),
  123--134 (2006)

\bibitem{izadinia2017im2cad}
Izadinia, H., Shan, Q., Seitz, S.M.: Im2cad. In: Proceedings of the IEEE
  conference on computer vision and pattern recognition. pp. 5134--5143 (2017)

\bibitem{karsch2011rendering}
Karsch, K., Hedau, V., Forsyth, D., Hoiem, D.: Rendering synthetic objects into
  legacy photographs. ACM Transactions on graphics (TOG)  \textbf{30}(6),
  1--12 (2011)

\bibitem{kawai2015diminished}
Kawai, N., Sato, T., Yokoya, N.: Diminished reality based on image inpainting
  considering background geometry. IEEE transactions on visualization and
  computer graphics  \textbf{22}(3),  1236--1247 (2015)

\bibitem{kulshreshtha2022layout}
Kulshreshtha, P., Lianos, N., Pugh, B., Jiddi, S.: Layout aware inpainting for
  automated furniture removal in indoor scenes. In: 2022 IEEE International
  Symposium on Mixed and Augmented Reality Adjunct (ISMAR-Adjunct). pp.
  839--844. IEEE (2022)

\bibitem{legendre2019deeplight}
LeGendre, C., Ma, W.C., Fyffe, G., Flynn, J., Charbonnel, L., Busch, J.,
  Debevec, P.: Deeplight: Learning illumination for unconstrained mobile mixed
  reality. In: Proceedings of the IEEE/CVF Conference on Computer Vision and
  Pattern Recognition. pp. 5918--5928 (2019)

\bibitem{li2020inverse}
Li, Z., Shafiei, M., Ramamoorthi, R., Sunkavalli, K., Chandraker, M.: Inverse
  rendering for complex indoor scenes: Shape, spatially-varying lighting and
  svbrdf from a single image. In: Proceedings of the IEEE/CVF Conference on
  Computer Vision and Pattern Recognition. pp. 2475--2484 (2020)

\bibitem{li2022physically}
Li, Z., Shi, J., Bi, S., Zhu, R., Sunkavalli, K., Ha{\v{s}}an, M., Xu, Z.,
  Ramamoorthi, R., Chandraker, M.: Physically-based editing of indoor scene
  lighting from a single image. In: European Conference on Computer Vision. pp.
  555--572. Springer (2022)

\bibitem{liu2020single}
Liu, Y.L., Lai, W.S., Chen, Y.S., Kao, Y.L., Yang, M.H., Chuang, Y.Y., Huang,
  J.B.: Single-image hdr reconstruction by learning to reverse the camera
  pipeline. In: Proceedings of the IEEE/CVF Conference on Computer Vision and
  Pattern Recognition. pp. 1651--1660 (2020)

\bibitem{mitsunaga1999radiometric}
Mitsunaga, T., Nayar, S.K.: Radiometric self calibration. In: Proceedings. 1999
  IEEE computer society conference on computer vision and pattern recognition
  (Cat. No PR00149). vol.~1, pp. 374--380. IEEE (1999)

\bibitem{moeck2007accuracy}
Moeck, M.: Accuracy of luminance maps obtained from high dynamic range images.
  Leukos  \textbf{4}(2),  99--112 (2007)

\bibitem{nie2020total3dunderstanding}
Nie, Y., Han, X., Guo, S., Zheng, Y., Chang, J., Zhang, J.J.:
  Total3dunderstanding: Joint layout, object pose and mesh reconstruction for
  indoor scenes from a single image. In: Proceedings of the IEEE/CVF Conference
  on Computer Vision and Pattern Recognition. pp. 55--64 (2020)

\bibitem{reinhard2010high}
Reinhard, E., Heidrich, W., Debevec, P., Pattanaik, S., Ward, G., Myszkowski,
  K.: High dynamic range imaging: acquisition, display, and image-based
  lighting. Morgan Kaufmann (2010)

\bibitem{srinivasan2020lighthouse}
Srinivasan, P.P., Mildenhall, B., Tancik, M., Barron, J.T., Tucker, R.,
  Snavely, N.: Lighthouse: Predicting lighting volumes for spatially-coherent
  illumination. In: Proceedings of the IEEE/CVF Conference on Computer Vision
  and Pattern Recognition. pp. 8080--8089 (2020)

\bibitem{stumpfel2006direct}
Stumpfel, J., Jones, A., Wenger, A., Tchou, C., Hawkins, T., Debevec, P.:
  Direct hdr capture of the sun and sky. In: SIGGRAPH 2006 Courses, pp. 5--es.
  ACM (2006)

\bibitem{suvorov2022resolution}
Suvorov, R., Logacheva, E., Mashikhin, A., Remizova, A., Ashukha, A.,
  Silvestrov, A., Kong, N., Goka, H., Park, K., Lempitsky, V.:
  Resolution-robust large mask inpainting with fourier convolutions. In:
  Proceedings of the IEEE/CVF winter conference on applications of computer
  vision. pp. 2149--2159 (2022)

\bibitem{wang2018self}
Wang, F.E., Hu, H.N., Cheng, H.T., Lin, J.T., Yang, S.T., Shih, M.L., Chu,
  H.K., Sun, M.: Self-supervised learning of depth and camera motion from 360
  videos. In: Asian Conference on Computer Vision. pp. 53--68. Springer (2018)

\bibitem{BiFuse20}
Wang, F.E., Yeh, Y.H., Sun, M., Chiu, W.C., Tsai, Y.H.: Bifuse: Monocular 360
  depth estimation via bi-projection fusion. In: The IEEE/CVF Conference on
  Computer Vision and Pattern Recognition (CVPR) (June 2020)

\bibitem{wang2021led2}
Wang, F.E., Yeh, Y.H., Sun, M., Chiu, W.C., Tsai, Y.H.: Led2-net: Monocular
  360deg layout estimation via differentiable depth rendering. In: Proceedings
  of the IEEE/CVF Conference on Computer Vision and Pattern Recognition. pp.
  12956--12965 (2021)

\bibitem{xiao2012recognizing}
Xiao, J., Ehinger, K.A., Oliva, A., Torralba, A.: Recognizing scene viewpoint
  using panoramic place representation. In: 2012 IEEE Conference on Computer
  Vision and Pattern Recognition. pp. 2695--2702. IEEE (2012)

\bibitem{yang2022automated}
Yang, B., Jiang, T., Wu, W., Zhou, Y., Dai, L.: Automated semantics and
  topology representation of residential-building space using floor-plan raster
  maps. IEEE Journal of Selected Topics in Applied Earth Observations and
  Remote Sensing  \textbf{15},  7809--7825 (2022)

\bibitem{yang2019dula}
Yang, S.T., Wang, F.E., Peng, C.H., Wonka, P., Sun, M., Chu, H.K.: Dula-net: A
  dual-projection network for estimating room layouts from a single rgb
  panorama. In: Proceedings of the IEEE/CVF Conference on Computer Vision and
  Pattern Recognition. pp. 3363--3372 (2019)

\bibitem{yeh2022photoscene}
Yeh, Y.Y., Li, Z., Hold-Geoffroy, Y., Zhu, R., Xu, Z., Ha{\v{s}}an, M.,
  Sunkavalli, K., Chandraker, M.: Photoscene: Photorealistic material and
  lighting transfer for indoor scenes. In: Proceedings of the IEEE/CVF
  Conference on Computer Vision and Pattern Recognition. pp. 18562--18571
  (2022)

\bibitem{zeng2019deep}
Zeng, Z., Li, X., Yu, Y.K., Fu, C.W.: Deep floor plan recognition using a
  multi-task network with room-boundary-guided attention. In: Proceedings of
  the IEEE/CVF International Conference on Computer Vision. pp. 9096--9104
  (2019)

\bibitem{zhang2019orientation}
Zhang, C., Liwicki, S., Smith, W., Cipolla, R.: Orientation-aware semantic
  segmentation on icosahedron spheres. In: Proceedings of the IEEE/CVF
  International Conference on Computer Vision. pp. 3533--3541 (2019)

\bibitem{zhang2016emptying}
Zhang, E., Cohen, M.F., Curless, B.: Emptying, refurnishing, and relighting
  indoor spaces. ACM Transactions on Graphics (TOG)  \textbf{35}(6),  1--14
  (2016)

\bibitem{zhang2021no}
Zhang, E., Martin-Brualla, R., Kontkanen, J., Curless, B.L.: No shadow left
  behind: Removing objects and their shadows using approximate lighting and
  geometry. In: Proceedings of the IEEE/CVF Conference on Computer Vision and
  Pattern Recognition. pp. 16397--16406 (2021)

\bibitem{zhang2014panocontext}
Zhang, Y., Song, S., Tan, P., Xiao, J.: Panocontext: A whole-room 3d context
  model for panoramic scene understanding. In: Computer Vision--ECCV 2014: 13th
  European Conference, Zurich, Switzerland, September 6-12, 2014, Proceedings,
  Part VI 13. pp. 668--686. Springer (2014)

\bibitem{zhi2022semantically}
Zhi, T., Chen, B., Boyadzhiev, I., Kang, S.B., Hebert, M., Narasimhan, S.G.:
  Semantically supervised appearance decomposition for virtual staging from a
  single panorama. ACM Transactions on Graphics (TOG)  \textbf{41}(4),  1--15
  (2022)

\bibitem{zhou2018semantic}
Zhou, B., Zhao, H., Puig, X., Xiao, T., Fidler, S., Barriuso, A., Torralba, A.:
  Semantic understanding of scenes through the ade20k dataset. International
  Journal on Computer Vision  (2018)

\bibitem{zou2018layoutnet}
Zou, C., Colburn, A., Shan, Q., Hoiem, D.: Layoutnet: Reconstructing the 3d
  room layout from a single rgb image. In: Proceedings of the IEEE conference
  on computer vision and pattern recognition. pp. 2051--2059 (2018)

\end{thebibliography}
\end{document}